\documentclass{llncs}
\usepackage{makeidx}  
\usepackage{url}      
\usepackage{subfigure}
\usepackage[linesnumbered,ruled,vlined]{algorithm2e}
\usepackage{amsfonts}
\usepackage{amsmath}
\usepackage{cancel}
\usepackage{graphicx}
\usepackage{color}
\usepackage{listings}
\usepackage{hyperref}
\hypersetup{
    colorlinks=true,
    linkcolor=blue,
    filecolor=magenta,      
    urlcolor=blue,
}
 
\usepackage{tikz}
\usepackage{ifthen}
\usepackage{xcolor}
\usepackage{multirow}
\usetikzlibrary{shadows.blur}
\usepackage{booktabs}
\usepackage{xspace}

\newlength{\forkmeoffset}
\definecolor{forkmebg}{HTML}{CC0000}
\definecolor{forkmefg}{HTML}{EEEEEE}

\DeclareMathOperator*{\argmax}{arg\,max}

\usepackage{graphicx}
\graphicspath{{pics/}{figs/}}

\begin{document}

\frontmatter          
\pagestyle{headings}  
\mainmatter              

\title{Searching Learning Strategy with Reinforcement Learning for 3D Medical Image Segmentation}
\titlerunning{Searching Learning Strategy with Reinforcement Learning for 3D Medical Image Segmentation}

\author{
 Dong Yang,
 Holger Roth,
 Ziyue Xu,
 Fausto Milletari,
 Ling Zhang,\\
 Daguang Xu\thanks{Corresponding author, \href{mailto:daguangx@nvidia.com}{daguangx@nvidia.com}.}
}

\institute{
 NVIDIA
}
\authorrunning{***}

\maketitle              

\begin{abstract}
Deep neural network (DNN) based approaches have been widely investigated and deployed in medical image analysis. For example, fully convolutional neural networks (FCN) achieve the state-of-the-art performance in several applications of 2D/3D medical image segmentation. Even the baseline neural network models (U-Net, V-Net, etc.) have been proven to be very effective and efficient when the training process is set up properly. Nevertheless, to fully exploit the potentials of neural networks, we propose an automated searching approach for the optimal training strategy with reinforcement learning. The proposed approach can be utilized for tuning hyper-parameters, and selecting necessary data augmentation with certain probabilities. The proposed approach is validated on several tasks of 3D medical image segmentation. The performance of the baseline model is boosted after searching, and it can achieve comparable accuracy to other manually-tuned state-of-the-art segmentation approaches.
\end{abstract}
\section{Introduction}
\label{sec:intro}
Medical image segmentation plays an important role in research and clinical practice and is necessary for tasks such as disease diagnosis, treatment planning, guidance, and surgery.
Researchers have been developing various automated and semi-automated approaches for 2D/3D medical image segmentation.
Among the prevailing approaches, deep neural networks (DNNs) have been successfully deployed for image/volume segmentation during past few years~\cite{choy2018current}.
Deep neural networks are capable to not only achieve state-of-the-art accuracy at inference but also to deliver results in a quick and efficient manner due to readily available GPU-accelerated computing routines.
So far, many baseline neural network models \cite{ronneberger2015u,milletari2016v,liu20183d} have been created and validated for various segmentation applications.
However, training such models requires careful design of the work-flow, and setup of data augmentation, learning rate, loss functions, optimizer and so on.
To achieve state-of-the-art performance, the model hyper-parameters need to be well-tuned, based either on extensive experimentation and grid parameter search or heuristics stemming from specific domain knowledge and expertise.

Recent works indicate that the full potential of current state-of-the-art network models may not yet be well-explored.
For instance, the winning solution of the \textit{Medical Decathlon Challenge} \cite{msd2018,isensee2018nnu} (consisting of ten 3D image segmentation tasks) is using ensembles of 2D/3D U-Net only, and elaborate engineering designs.
The argument raised by such work is that the potentials of the successful baseline models may be neglected.
This argument cannot be easily confirmed since the theoretical explanation of deep neural network has not been well-established. 
Therefore, although the current research trend is to develop elaborate and powerful 3D segmentation network models (within GPU memory limit), it is also very important to pay attentions to the details of model training.

\textit{Automatic Machine Learning} (AutoML) has been recently proposed to automatically search the best learning approaches and minimize human interaction at the same time.
Different approaches \cite{zoph2016neural,liu2018progressive,liu2018darts,zoph2018learning,pham2018efficient,liu2019auto} have been introduced in computer vision to search for the best neural network architecture for image analysis and scene understanding tasks.
Unlike specifically designed networks (e.g. ResNet~\cite{he2016deep}), the often peculiar neural architectures resulting from the automatic search process can achieve state-of-the-art performance for tasks at hand.

Since 3D segmentation is very expensive to train, efficient 3D architecture search is extremely difficult to attain. Instead, a more feasible task is represented by hyper-parameter searching which still plays a central role on the test-time performance.
In this work, we propose a reinforcement learning-based approach to search the best training strategy of deep neural networks for a specific 3D medical image segmentation task.
Training strategies include the learning rate, data augmentation strategies, data pre-processing, etc. 
In the proposed framework, an additional recurrent neural network (RNN) - the controller - is trained to generate hyper-parameters of the training strategies.
The reward signal supplied during training to our RL-based controller is the validation accuracy of segmentation network.
The RNN is trained with the reward and observation (the previous set of training strategies).
Finally, the best strategy is generated once the searching process is done.

\section{Related Work}
\label{sec:related}
In machine learning, the hyper-parameter optimization has been studies for years, and several approaches have been developed such as grid search, Bayesian optimization, random search and so on~\cite{bergstra2011algorithms}.
The main idea of grid search and random search is using brute force to enumerate possible hyper-parameter combinations in order to determine the one with the best validation accuracy.
They can be naturally implemented in parallel.
However, in practice, such approaches only work well within a low-dimensional searching space and they becomes extremely impractical once the searching space is large with a high dimension.
Bayesian optimization like \textit{Gaussian Processes} (GP) normally generate better results in fewer steps comparing with the ``brute-force'' approaches because the feedback from each training process is actually used for updating posterior functions, and generating the next parameter settings for searching.
On the other hand, it requires the definition of a prior function to describe the behaviors of the objective, and the final performance relies heavily on the choice of this prior.

Recently, researchers proposed reinforcement learning (RL) based approaches for neural architecture searching~\cite{zoph2016neural,zoph2018learning,pham2018efficient}.
In principle, a RNN-based agent/policy collects the information (reward, state) from the environment, update the weights within itself, and creates the next potential neural architectures for validation.
The searching objectives are the parameters of the convolutional kernels, and how they are connected one-by-one.
The validation output is utilized as the reward to update the agent/policy.
The RL related approaches fit such scenario since there is no ground truth for the neural architectures with the best validation performance.
In order to avoid the request of huge amounts of GPU hours, researchers also investigated more efficient ways to conduct neural architecture search.
The progressive neural architecture search introduced a new way to construct new neural architecture on the fly during training~\cite{liu2018progressive}.
Furthermore, the differentiable architecture search presented a searching strategy through learning a weighted sum of potential components, and finalizing a discrete architecture with $\argmax$ operations.
However, there might be an unexpected gap between the ``continuous'' and discretized architectures in specific applications.
In addition, the differentiable architecture search requires to load all possible neural components during training, which potentially takes a lot of GPU memory, especially in 3D image processing.
Alternatively, RL based approaches can be applied to the tasks of optimizing other parts of neural network model training, such as data augmentation policies~\cite{cubuk2018autoaugment} and design of loss functions~\cite{xu2018autoloss} in order to bypass some limitations of differentiable architecture search.

\section{Methodology}
\label{sec:method}
In this section, we firstly introduce the definition of the searching space in our framework and then describe the RNN controller, and how the searching procedure is achieved using RL.
The proposed approach is inspired by the work ``auto-augment''~\cite{cubuk2018autoaugment}.
We expand its original idea and apply our extended version to 3D medical image segmentation.
\subsection{Searching Space Definition}
The upper-bound performance of a machine learning model is always directly limited by the hyper-parameter setup during training.
For example, the learning rate for weight updates is critical for achieving decent performance in most deep learning applications.
The conventional way to determine hyper-parameters is to use domain knowledge and necessary heuristics.
However, some of the hyper-parameters are often highly related to the dataset itself, which may not be easily understood by humans.
Thus, we propose to automatically search within certain hyper-parameter spaces, to ease the burden for setting those numbers.

In our setting, given a parameter $\lambda$, the maximum value $\lambda_{max}$ and minimum value $\lambda_{min}$ are required.
Then the range between $\lambda_{max}$ and $\lambda_{min}$ is mapped to $\left[0,1\right]$.
The floating number $\lambda^{*}$ is the searching target with the optimal performance.
For a set of hyper-parameters $\Lambda=\left\{\lambda_{1},\lambda_{2},...\right\}$, the searching objective is $\Lambda^{*}=\left\{\lambda_{1}^{*},\lambda_{2}^{*},...\right\}$.
Each $\lambda_{i}^{*}$ may not be optimal for that specific parameter, since greedy-type searching algorithms may lead to sub-optimal performance.
Therefore, it is necessary to search for all the hyper-parameter jointly.

Firstly, we consider the parameters for data augmentation, which is an important component for training neural networks in 3D medical image segmentation as it increases the robustness of the models and avoids overfitting~\cite{isensee2018nnu}.
Augmentation includes image sharpening, image smoothing, adding Gaussian noise, contrast adjustment, and random shift of intensity range, etc.
We assign a probability value $p_{i}$ to each augmentation approach to determine how likely the corresponding augmentation will occur.
During training, a random number $r_{i}\in \left ( 0,1 \right )$ for the $i$th augmentation approach is generated at each iteration.
If $r_{i}\geq a$, the augmentation is executed; if $r_{i}<a$, this augmentation will not be conducted.
In this way, we can have information about the relative importance between different augmentation approaches according to the specific dataset or application.
Secondly, we found the learning rate $\alpha$ is also critical for medical image segmentation.
Sometimes, large network models favor a large $\alpha$ for activation, and small datasets prefer small $\alpha$. Once $\alpha_{max}$ and $\alpha_{min}$ are determined, the searching range of $\alpha$ is set and mapped to $\left ( 0,1 \right )$.
Similar treatment can be applied to any possible hyperparameters in the training process for optimization.
Moreover, unlike other approaches, we search for the optimal hyper-parameters in the high-dimensional continuous space instead of discrete space.

\begin{algorithm}[H]
\label{alg:the_alg}
\SetAlgoLined
\KwResult{Optimal training strategy $C^*$, given a dataset $D$}
 Set $\mathrm{MAX\_EPOCH}=1000,\mathrm{EPOCH}=0$\;
 Random initialization $C_1,C_2,C_3,\cdots$\;
 Launch training jobs $T\left(C_1\right),T\left(C_2\right),T\left(C_3\right),\cdots$\;
 \While{$\mathrm{EPOCH} < \mathrm{MAX\_EPOCH}$}{
  Collect validation accuracy $V_i=T\left(C_i\right)$ from a finished job\;
  Update weights of RNN controller\;
  Generate a new training strategy $C_j$\;
  Launch a training job $T\left(C_j\right)$\; 
  $\mathrm{EPOCH}=\mathrm{EPOCH}+1$\;
 }
 \caption{RL based training strategy searching}
\end{algorithm}
\subsection{RL based Searching Approach}
Because there is no ground truth for the optimal validation accuracy, RL fits the scenario to derive the optimal training strategy/configuration $C$ given specific dataset $D$.
Our searching approach is shown in Algorithm~\ref{alg:the_alg}.
During the process, an RNN-based job controller $H$ is created for communicating with different launched jobs.
In the beginning, $H$ launches $n$ training jobs with randomly initialized training configurations $C_i$.
$C_i$ can be defined as a vector, and each element is sampled from one dimension of the aforementioned searching space.
Also, $C_i$ is sufficient to accomplish each training job $T$.
After training, the validation accuracy $V=T\left(C_i\right)$ is returned to $H$ for updating the weights of the RNN in controller, and generating a new strategy $C$ for future training epochs.

\begin{figure}[!t]
    \centering
    \includegraphics[width=\columnwidth]{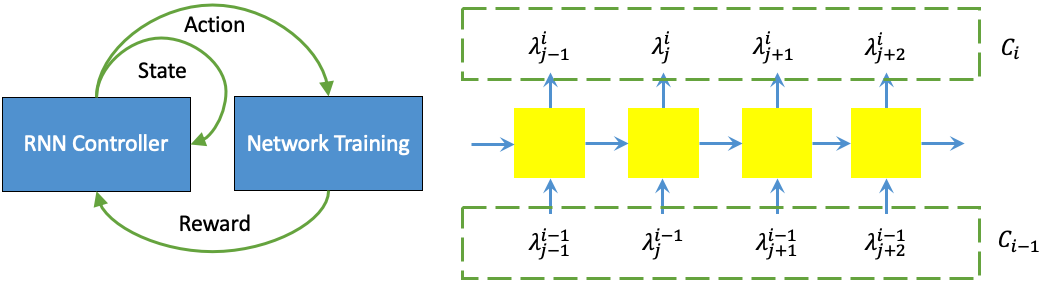}
    \caption{Left: the communication between training jobs and RNN controller $H$. The RNN controller provides the action and state for next step, and network training produces a reward for $H$. Right: previous training strategy $C_{i-1}$ is the input of $H$, and output is the next strategy $C_i$ after updating the weights.}
    \label{fig:framework}
\end{figure}

Our framework is shown in Figure~\ref{fig:framework}.
For the RL setting, the reward is the validation accuracy, the action is the newly generated $C_i$, environment observation/state is $C_{i-1}$ from the last step, and the policy is the RNN job controller $H$.
$H$ is a basic recurrent unit with one hidden layer.
And the input nodes (observation) and output nodes (action) of $H$ share the same quantity.
Each output node produces two-channel outputs after softmax activation.
Then the first channel of the output is fed to the next step as action after mapping back to the original searching space.
The \textit{Proximal Policy Optimization} (PPO) is adopted to train the RNN cells in $H$~\cite{schulman2017proximal}.
The loss function is as follows.
\begin{equation}
    \theta\leftarrow \theta+\gamma r\nabla_{\theta}\mathrm{ln}H \left ( C_i |C_{i-1},\theta\right )
\end{equation}
Here, $\theta$ represents the weights in RNN.
During training, the reward $r$ is utilized to update the weights using gradient back-propagation.
To train the RNN controller, we use RMSprop as the optimizer with a learning rate $\gamma$ of 0.1.

\section{Experimental Evaluation}
\label{sec:experiments}
\textbf{Datasets}~The medical decathlon challenge (MSD) provides ten different tasks on 3D CT/MR image segmentation~\cite{msd2018}.
Datasets of task02 (left atrium segmentation), task06 (lung tumor segmentation), task07 (pancreas and tumor segmentation), and task09 (spleen segmentation) are used with our own random split for training/validation.
For task02, 16 MR volumes for training, 4 for validation.
For task06, 50 CT volumes for training, 13 for validation.
For task07, 224 CT volumes for training, 57 for validation.
And for task09, 32 CT volumes for training, 9 for validation.
We re-sample both the images and labels into the isotropic resolution 1.0$mm$.
The voxel intensities of the images are normalized to the range $\left[0,1\right]$ according to the following input ranges: 5th and 95th percentile of overall voxel intensities for MRI and -1000 and 1000 Hounsfield units for CT.

\noindent\textbf{Implementation}~Our baseline model follows the work the 2D-3D hybrid network proposed in~\cite{liu20183d}, but without the \textit{PSP} component.
The pre-trained ResNet-50 (on ImageNet) possesses a powerful capability for feature extraction as the encoder.
And the 3D decoder network with \textit{DenseBlock} provides smooth 3D predictions.
The input of the network are $96\times 96\times 96$ patches, randomly cropped from the re-sampled images during training.
Meanwhile, the validation step follows the scanning window scheme with a small overlap (one quarter of a patch).
By default, all training jobs use the Adam optimizer, and the Dice loss is used for gradient computing~\cite{milletari2016v}.
The validation accuracy is measured with the Dice's score after scanning window.
Our method is implemented with TensorFlow and trained on NVIDIA V100 GPUs with 16 GB memory.

Firstly, we compare the proposed approach with the local searching approaches, shown in Table~\ref{tab:fine-tune}.
The hill climbing algorithm is a classical greedy search technique.
It is more efficient than grid or random search if the initial position is properly set.
We implemented two versions: discrete searching and continuous searching.
The discrete version assumes the searching space is discrete with fixed dimension, and step size at each move is fixed.
In our setting, the dimension of the searching space for one parameter is 100, and the step size is 0.01.
At each move, the target value cannot go above $1.0$ or below $0.0$.
The continuous version uses an adaptive step.
If the moving direction improves the value, the step size will be multiplied by $1.1$, otherwise, it will be divided by $1.1$.
The searching space has to be positive for the continuous version.
Both versions stop when the local minimum is reached after convergence.

To save searching time, we start the searching process from a pre-trained model trained after 500 epochs without any augmentation or parameter searching.
In Table~\ref{tab:fine-tune}, ``no augmentation'' indicates the performance of the pre-trained model.
In our proposed approach, each job fine-tunes the pre-trained model with 200 epochs with its training strategy.
To make a fair comparison, the initial status before searching is set to 0.5 for all parameters to search (the searching space for learning rate is $\left [ 0.01, 0.0001 \right ]$).
After searching, our approach outperforms other baseline approaches according to the overall Dice's score in the validation dataset in the tested applications.
The maximum epoch number in our approach is set as 400, which takes circa 24 hours to finish searching with 32 GPUs running in parallel.
The baseline approach normally takes longer time because of the large searching space and small step size being employed.

\begin{table}
    \begin{center}
    \begin{tabular}{*4c}
    \toprule
    \multicolumn{2}{c}{MSD task02} & \multicolumn{2}{c}{MSD task09}\\
    \midrule
    Method & Validation Acc. & Method & Validation Acc.\\
    \midrule
    No Augmentation & 0.88 & No Augmentation & 0.87\\
    Discrete Hill Climbing & 0.90 & Discrete Hill Climbing & 0.90\\
    Continuous Hill Climbing & 0.90 & Continuous Hill Climbing  & 0.92\\
    Proposed Approach & \textbf{0.92} & Proposed Approach & \textbf{0.92}\\
    \bottomrule
    \end{tabular}
    \end{center}
    \caption{Performance comparison with baseline approaches and our proposed approach. The validation accuracy is the overall average Dice's score among different subjects and classes.}
    \label{tab:fine-tune}
\end{table}

Secondly, we conduct another experiment with three different datasets to validate the effectiveness of our proposed approach for the models trained from scratch.
``Routine'' means all the searched parameters are fixed as 0.5 (learning rate is set as $0.0001$), and the model is trained from scratch.
Our proposed approach searches from the same setting as ``routine''.
The maximum epoch number is 100 in our approach.
And the entire searching procedure takes about 48 hours with 50 GPUs.
All training jobs are trained with 800 epochs.
We can see that the models after searching work better, and the fixed parameter is clearly not optimal for these applications.
From the resulting parameters after the search, we can see clearly that each application has a preference for different augmentations or hyper-parameters.
For instance, the task09 is the spleen segmentation in CT.
According to CT imaging quality, the training strategy containing a random intensity scale shift would perform better.
The similar conclusion can be achieved from other CT datasets (e.g. task06, task07).
For MRI segmentation, the image sharpening is preferable as can be seen from the resulting training strategy.
The reason might be that the MRI quality varies a lot, and sharpening operation can strengthen the region-of-interest, especially the boundary regions.

\begin{table}
    \begin{center}
    \begin{tabular}{*6c}
    \toprule
    \multicolumn{2}{c}{MSD task06} & \multicolumn{2}{c}{MSD task07} & \multicolumn{2}{c}{MSD task09}\\
    \midrule
    Method & Validation Acc. & Method & Validation Acc. & Method & Validation Acc.\\
    \midrule
    Routine & 0.383 & Routine & 0.491 & Routine & 0.957\\
    Proposed & \textbf{0.449} & Proposed & \textbf{0.519} & Proposed & \textbf{0.960}\\
    \bottomrule
    \end{tabular}
    \end{center}
    \caption{Performance comparison of models training from scratch with or without using our proposed approach. The validation accuracy is the overall average Dice's score among different subjects and classes.}
    \label{tab:scratch}
\end{table}

The same task, task09, is used in both, the first and second experiment.
From the Table~\ref{tab:fine-tune} and Table~\ref{tab:scratch}, we can see training from scratch with augmentation could achieve a higher Dice's score compared with the one fine-tuned from a ``no-augmentation'' model.
This suggests that the found data augmentation strategy is effective when applied to training from scratch.

\section{Conclusions}
\label{sec:conclusions}
In this paper, we proposed a RL-based searching approach to optimize the training strategy for 3D medical image segmentation.
The proposed approach has been validated on several segmentation tasks with clear effectiveness.
It also possesses large potentials to be applied for general machine learning problems.
For example, the heuristic parts of any learning algorithm can be easily determined after optimization or searching, given a specific medical imaging application.
Moreover, extending the single-value reward function to a multi-dimensional reward function could be studied as the future direction.
%

\begin{thebibliography}{10}
\providecommand{\url}[1]{\texttt{#1}}
\providecommand{\urlprefix}{URL }

\bibitem{msd2018}
Medical decathlon challenge (2018), \url{http://medicaldecathlon.com}

\bibitem{bergstra2011algorithms}
Bergstra, J.S., Bardenet, R., Bengio, Y., K{\'e}gl, B.: Algorithms for
  hyper-parameter optimization. In: Advances in neural information processing
  systems. pp. 2546--2554 (2011)

\bibitem{choy2018current}
Choy, G., Khalilzadeh, O., Michalski, M., Do, S., Samir, A.E., Pianykh, O.S., et al.: Current
  applications and future impact of machine learning in radiology. Radiology
  288(2),  318--328 (2018), doi: \url{10.1148/radiol.2018171820}.

\bibitem{cubuk2018autoaugment}
Cubuk, E.D., Zoph, B., Mane, D., Vasudevan, V., Le, Q.V.: Autoaugment: Learning
  augmentation policies from data. arXiv preprint arXiv:1805.09501  (2018)

\bibitem{he2016deep}
He, K., Zhang, X., Ren, S., Sun, J.: Deep residual learning for image
  recognition. In: Proceedings of the IEEE conference on computer vision and
  pattern recognition. pp. 770--778 (2016), \url{doi: 10.1109/cvpr.2016.90}.

\bibitem{isensee2018nnu}
Isensee, F., Petersen, J., Klein, A., Zimmerer, D., Jaeger, P.F., Kohl, S., et al.: nnu-net:
  Self-adapting framework for u-net-based medical image segmentation. arXiv
  preprint arXiv:1809.10486 (2018)

\bibitem{liu2019auto}
Liu, C., Chen, L.C., Schroff, F., Adam, H., Hua, W., Yuille, A., Fei-Fei, L.:
  Auto-deeplab: Hierarchical neural architecture search for semantic image
  segmentation. arXiv preprint arXiv:1901.02985 (2019)

\bibitem{liu2018progressive}
Liu, C., Zoph, B., Neumann, M., Shlens, J., Hua, W., Li, L.J., et al.: Progressive neural architecture search.
  In: Proceedings of the European Conference on Computer Vision (ECCV). pp.
  19--34 (2018)

\bibitem{liu2018darts}
Liu, H., Simonyan, K., Yang, Y.: Darts: Differentiable architecture search.
  arXiv preprint arXiv:1806.09055 (2018)

\bibitem{liu20183d}
Liu, S., Xu, D., Zhou, S.K., Pauly, O., Grbic, S., Mertelmeier, et al.: 3d anisotropic hybrid network:
  Transferring convolutional features from 2d images to 3d anisotropic volumes.
  In: International Conference on Medical Image Computing and Computer-Assisted
  Intervention. pp. 851--858. Springer (2018), doi: \url{10.1007/978-3-030-00934-2-94}.

\bibitem{milletari2016v}
Milletari, F., Navab, N., Ahmadi, S.A.: V-net: Fully convolutional neural
  networks for volumetric medical image segmentation. In: 2016 Fourth
  International Conference on 3D Vision (3DV). pp. 565--571. IEEE (2016), doi: 10.1109/3dv.2016.79.

\bibitem{pham2018efficient}
Pham, H., Guan, M., Zoph, B., Le, Q., Dean, J.: Efficient neural architecture
  search via parameter sharing. In: International Conference on Machine
  Learning. pp. 4092--4101 (2018)

\bibitem{ronneberger2015u}
Ronneberger, O., Fischer, P., Brox, T.: U-net: Convolutional networks for
  biomedical image segmentation. In: International Conference on Medical image
  computing and computer-assisted intervention. pp. 234--241. Springer (2015), doi: \url{10.1007/978-3-319-24574-4-28}.

\bibitem{schulman2017proximal}
Schulman, J., Wolski, F., Dhariwal, P., Radford, A., Klimov, O.: Proximal
  policy optimization algorithms. arXiv preprint arXiv:1707.06347 (2017)

\bibitem{xu2018autoloss}
Xu, H., Zhang, H., Hu, Z., Liang, X., Salakhutdinov, R., Xing, E.: Autoloss:
  Learning discrete schedules for alternate optimization. arXiv preprint
  arXiv:1810.02442 (2018)

\bibitem{zoph2016neural}
Zoph, B., Le, Q.V.: Neural architecture search with reinforcement learning.
  arXiv preprint arXiv:1611.01578 (2016)

\bibitem{zoph2018learning}
Zoph, B., Vasudevan, V., Shlens, J., Le, Q.V.: Learning transferable
  architectures for scalable image recognition. In: Proceedings of the IEEE
  conference on computer vision and pattern recognition. pp. 8697--8710 (2018), doi: \url{10.1109/cvpr.2018.00907}.

\end{thebibliography}
\bibliographystyle{splncs03}

\end{document}